\documentclass[12pt,letterpaper]{article}

\usepackage{times}
\usepackage{latexsym}
\usepackage{amsmath}
\usepackage{multirow}
\usepackage{subcaption}

\newcommand{\captionfonts}{\normalsize}

\makeatletter  
\long\def\@makecaption#1#2{%
  \vskip\abovecaptionskip
  \sbox\@tempboxa{{\captionfonts #1: #2}}%
  \ifdim \wd\@tempboxa >\hsize
    {\captionfonts #1: #2\par}
  \else
    \hbox to\hsize{\hfil\box\@tempboxa\hfil}%
  \fi
  \vskip\belowcaptionskip}
\makeatother   

\usepackage{natbib}  
\newcommand{\parencite}{\citep}
\newcommand{\textcite}{\citet}
\renewcommand{\cite}{\citealt}

\usepackage[autostyle]{csquotes}

\usepackage{url}

\usepackage[T1]{fontenc}    
\usepackage{url}            
\usepackage{booktabs}       
\usepackage{amsfonts}       
\usepackage{nicefrac}       
\usepackage{microtype}      
\usepackage{amssymb}
\usepackage{units}
\usepackage{graphics}
\usepackage{graphicx}
\usepackage{floatrow} 
\usepackage{algorithm}
\usepackage{algorithmicx}
\usepackage[noend]{algpseudocode}
\usepackage{color}   
\usepackage{alltt}
\usepackage{mathtools}
\usepackage{amsfonts}
\usepackage{adjustbox}
\usepackage{todonotes}
\usepackage{subcaption}

\usepackage{enumitem}

\algnewcommand{\LineComment}[1]{\State \(//\) #1}
\algnewcommand{\IfThenElse}[3]{
  \State \algorithmicif\ #1\ \algorithmicthen\ #2\ \algorithmicelse\ #3}

\begin{document}
\hspace{13.9cm}

\ \vspace{20mm}\\
{\LARGE Learning to Adapt by Minimizing Discrepancy}
\ \\
{\bf \large
Alexander G. Ororbia II$^{\displaystyle 1}$} \\
{\bf \large
Patrick Haffner$^{\displaystyle 3}$} \\
{\bf \large
David Reitter$^{\displaystyle 1}$} \\
{\bf \large
C. Lee Giles$^{\displaystyle 1}$} \\
{$^{\displaystyle 1}$The Pennsylvania State University}\\
{$^{\displaystyle 3}$Interactions, LLC.}\\

{\bf Keywords: adaptive learning, neural architecture} 

\thispagestyle{empty}

\begin{abstract}
We explore whether useful temporal neural generative models can be learned from sequential data without back-propagation through time. We investigate the viability of a more neurocognitively-grounded approach in the context of unsupervised generative modeling of sequences. Specifically, we build on the concept of predictive coding, which has gained influence in cognitive science \citep{rauss2013predictive,panichello2013predictive}, in a neural framework. To do so we develop a novel architecture, the Temporal Neural Coding Network, and its learning algorithm, Discrepancy Reduction. 
The underlying directed generative model is \emph{fully recurrent}, meaning that it employs structural feedback connections and temporal feedback connections, yielding information propagation cycles that create local learning signals. This facilitates a unified bottom-up and top-down approach for information transfer inside the architecture. Our proposed algorithm shows promise on the bouncing balls generative modeling problem. Further experiments could be conducted to explore the strengths and weaknesses of our approach. \footnote{An earlier version of the ideas in this paper was presented as a talk \citep{ororbia2017tncn} at the \emph{11th Annual Machine Learning Symposium}.}
\end{abstract}

\section{Introduction}
\label{intro}
For many problems, back-propagation of errors, or the application of reverse-mode differentiation to computation graphs, has been the primary algorithm of choice for conducting credit assignment in neural architectures. However, when neural architectures are made deeper, and thus more complex, the error gradients must pass backward through many layers. As a result of the additional multiplications, these gradients tend to either explode or vanish \citep{pascanu2013difficulty}. In order to keep the values of the gradients within reasonable magnitudes, we often require the layers to behave sufficiently linearly to prevent saturation of neuronal post-activities, which would yield zero gradient. It has been shown that this required linearity can lead to undesirable extrapolation effects, creating the well-known problem of adversarial samples \citep{szegedy2013intriguing,ororbia2017unifying}.
Furthermore, this linearity also hinders usage of other important, non-linear mechanisms, such as lateral competition and saturation.

From a biological perspective, back-propagation has received a great deal of criticism due to the implausibility of its implementation in the brain. Some of the key problems include: 1) the ``weight transport problem'', where the feedback weights must be the same as (the transpose of) the feedback weights, 2) the forward pass and the backward pass require different computations, and 3) the error gradients must be stored separately from the activations. To remedy the last two conditions, one could use a symmetrical network that is solely used for propagating errors (an error-propagation network). However, beyond the fact that two information pathways have been created, there is no known biological mechanism that allows the error network to know the weights of the feedforward network it is serving. The earlier described requirement of linearity also violates what we know about biological neurons, which interleave linear and non-linear operations. As argued by \textcite{bengio2015towards}, if the brain were to use feedback paths as implemented by back-propagation, it would require precise knowledge of the derivatives of the non-linear activation functions 
, which is not possible since not all neurons are the same.
Furthermore, discrete-valued or stochastic activations (such as sampling a Bernoulli or Categorical distribution) cannot be used, even though we know that real neurons communicate with spikes (binary values) and not by continuous values. 

More critically, back-propagation requires a global feedback path to carry error derivatives across the system. This is due to the nature of supervised learning systems--an objective is grounded in input/output space and the global pathway is used to relate how internal processing elements affect the target. One problem with this, especially when used to generatively model data, is that most of the learner's time is spent on surface-level properties of the data and not on extracting latent structure necessary for generalization. A good example is in speech processing, where the log likelihood cost used leads the model to focus mostly on acoustic details rather than higher-level linguistic features.\footnote{Yoshua Bengio, presentation at ReWork: Deep Learning Summit Montreal, 2017.}

This global feedback path stands in contrast to many theories of the brain \citep{grossberg1982does,rao1999predictive,huang2011predictive,clark2013whatever}, which posit that local computations occur at multiple levels of a (somewhat) hierarchical structure. However, if we were to violate the idea of a global feedback path, where would the targets then come from in order to create learning signals for the hidden processing elements? One thing is likely: we will no longer be able to rely on a loss function that operates primarily in the input space, which is at the core of supervised learning. This means that the learning approach we seek will require \emph{higher-level objectives}, or objectives that operate at various levels of latent space. 
More importantly, in designing higher-level objectives to create local targets, we can better encourage a neural system to find hidden/abstract structure in data. This type of objective directly connects to representation learning, better embodying one of the key assumptions behind unsupervised learning:  by observing a stream of data points, it is possible to derive the predictable systemic relationships between variables \footnote{These variables can be pixels in images of a video or the characters of a word in a sentence.} as well as relationships between these relationships, i.e., more complicated, abstract patterns. 
Higher-level patterns are what a representation learning system seeks to uncover--latent variables, or intermediate concepts and features, that capture useful statistical regularities of the world that the intelligent system is embedded in. To this end, in this paper, the intuition behind our learning algorithm is to measure and reduce the \emph{discrepancy}, or mismatch, between what representations a neural model can currently generate and representations that better describe the input.

Attempting to build models and algorithms that resolve some of the above criticisms might open the door to learning approaches that generalize better.
However, while many variations of back-propagation and alternatives have been proposed, most work has only shown their usefulness on static problems, typically on classification. However, we know that in the human brain, many active processes, including those related to vision and speech, take in sequences of input stimuli and attempt to build a dynamic mental model of the world \citep{rao1997dynamic}.  
Given this dynamic view of the brain, which constructs an implicit, abstract knowledge base that is representative of the structure of the observed environment \citep{reber1989implicit,destrebecqz2001can}, we design our approach with an eye toward stateful problems. From a machine learning perspective, this is important given the success of recurrent neural models sequential problems such as language modeling \citep{mikolov2010recurrent,ororbia2017diff}.
Critical to the success of recurrent models is the action of \emph{unrolling}, a mechanism that is clearly neurobiologically implausible. In order to implement back-propagation through time, one needs to unroll \footnote{Or unfold a recursively defined operation into an explicit chain of events.} the underlying computation graph over the length of a given input sequence, creating a longer global feedback path for error information to traverse. The brain, in contrast, is an incremental, adaptive process. As such, we investigate the viability of our model, which learns from sequences without any unrolling, and offer some promising evidence that our learning algorithm can match or outperform some powerful neural models that rely on back-propagation through time or advanced variations, such as neural variational inference \citep{mnih2014neural}. Notably, our proposed approach allows the learning of a directed generative model, which is important given the causal structure of the universe, without the need to correct for the imperfections of an approximate inference model.

The contributions of this article are the following: 
\begin{itemize}
  \item We propose the Temporal Neural Coding Network (TNCN), a temporal neural model, and its learning algorithm, \emph{Discrepancy Reduction}, for learning dynamically from sequential data. Our model incorporates some basic notions from predictive coding \citep{rao1999predictive} theories of the brain, notably lateral competition among neural variables.
  \item To create our model's unsupervised learning algorithm, we draw inspiration from random feedback alignment \citep{lillicrap2014random} and difference target propagation \citep{lee2015difference}. To create local targets for the model's higher-level objectives, we show that simple fixed projection functions can be used to create special error units that can generate local targets. 
    \item To evaluate our model and learning algorithm, we experiment with a video modeling problem and discover promising results with our learning approach.
\end{itemize}
This work can also be viewed as another contribution towards the long-term goal of finding more biologically plausible machine learning approaches to the credit assignment problem \cite{bengio2015towards}. Specifically, we offer a rather simple approach to implementing and training sequential predictive neural models without back-propagation through time.

\section{Motivation \& Related Work}
\label{motivation}
There has been a great deal of research in finding more biologically-plausible alternatives to back-propagation of errors. Classically, the online alternative to back-propagation of errors was real-time recurrent learning (RTRL, \cite{williams1989learning}), which employs forward-mode differentiation to compute gradients. However, this algorithm scales poorly, i.e., quadratically in the number of parameters. Some algorithms have been proposed to reduce the complexity of RTRL, including the \emph{NoBackTrack} \cite{ollivier2015training} and \emph{Unbiased Online Recurrent Learning} \cite{tallec2017unbiased} algorithms, but are noisy and slow down the learning procedure in trying to approximate the way back-propagation through time works in an online fashion.

The contrastive divergence recipe \citep{hinton2006training}, well-known for being the primary learning algorithm of restricted Boltzmann machines, and the Wake-Sleep algorithm \citep{hinton1995wake,bornschein2014reweighted} for training deeper Boltzmann-based architectures, were largely inspired by the role of sleep in human learning. However, these approaches to learning, which rely on Markov Chain Monte Carlo methods, suffer from a variety of problems including slow convergence to steady-state distributions due to poor mixing of modes and the constraint that the weights of the model must be symmetric. With some success, Boltzmann-based architectures have been applied to stateful problems \cite{taylor2007modeling,sutskever2009recurrent,boulanger2012modeling,mittelman2014structured}, but require hybridizing the Contrastive Divergence approach with back-propagation through time, incurring the limitations and criticisms of both algorithms. Other approaches inspired by Boltzmann-based learning (and energy-based learning in general) include the \emph{variational walkback} algorithm \citep{goyal2016variational} and \emph{equilibrium propagation} \citep{scellier2017equilibrium}. However, these algorithms have only been investigated on static modeling problems and it is not clear how one might extend them to stateful problems.

Learning deep Boltzmann machines can be quite difficult, requiring all sorts of tricks to make the process work well and efficiently \citep{salakhutdinov_efficient_2010,ororbia2015online}. In response, one line of work has taken on a variational inference scheme \citep{kingma2013auto,mnih2014neural,serban2017piecewise}, where we train an approximate inference machine parametrized by a neural network.  
While efficient in learning probabilistic models of data, the success of the generative model under the variational inference framework depends largely on how good the inference model is. Specifically, the inference model $q(\cdot)$ constrains the generative model $p(\cdot)$, and any deficiency in the inference model must be then be made up by the generative model. Instead we would like to reverse this scenario--the generative model adjusts itself, where generation can be used to prime the feedback mechanisms that will guide learning and adaptation. The proposed TNCN embodies this idea in the attempt to circumvent the need for approximate inference machinery.

Our algorithm is inspired by three different strands of research focused on finding viable alternatives to the biologically-implausible back-propagation of errors. 

\subsection{Random Feedback Alignment}
\label{fa}
Feedback alignment \citep{lillicrap2014random} and its variants \citep{nokland2016direct,baldi2016learning}, have shown that random feedback weights can deliver useful teaching signals. This random form of back-propagation has also been used to develop an event-driven variation of the learning rule suitable for neuromorphic implementations of neural networks \citep{neftci2017event}. More importantly, feedback alignment algorithms resolve the weight-transport problem, which has been one criticism of back-propagation before \citep{grossberg_resonance_1987,liao2016important}, by showing that coherent learning is possible with asymmetric forward and backward pathways. Rather, the error back-projection pathways need not be the transpose of the weights used to carry out forward propagation, and the learning process can instead be viewed as the alignment of feedforward weights with feedback weights. 

Feedback alignment, however, suffers from several problems: 1) during the alignment phase, a given layer cannot learn before the upper layers have approximately ``aligned'', 2) the procedure operates much like (supervised) greedy layerwise learning where each layer only learns something that is useful for a linear classifier but does not globally optimize or offer any coordination among the layers.\footnote{Personal communication, Yoshua Bengio.} The TNCN's learning algorithm deviates from feedback alignment in that it uses error feedback weights to create potentially better target representations instead of replacing the derivatives normally computed by back-propagation. Furthermore, the TNCN does not strive to learn by an algorithm that works approximately like back-propagation (as approaches like feedback alignment and contrastive Hebbian learning \citep{xie2003equivalence} do), which requires a global feedback path to conduct credit assignment.

\subsection{Recirculation \& Target Propagation}
\label{targ_prop}
Recirculation \citep{hinton1988learning,o1996biologically}, the predecessor to target propagation and originally proposed for a single hidden layer auto-encoder, uses the datum as the target value for reconstruction (which affects the decoder) and the initial encoded representation of the datum as the target for the encoder, which is computed after a second forward pass. One key requirement for recirculation is that the weights of the encoder and decoder are symmetric, however, the learning process encourages these weights to automatically self-align to approximate the symmetry. 

Difference target propagation \citep{lee2015difference} brings forth the idea that a learning signal might be created by instead calculating targets rather than loss gradients at each layer. This allows for the development of local learning rules, removing the need for a global error pathway to carry error derivatives across (and thus side-stepping the vanishing gradient problem). Furthermore, some connections can made between target propagation and  Spike-Time-Dependent-Plasticity \citep{andrew2003spiking}. The general approach in target propagation is that the feedback weights are trained to learn the inverse of the feedforward mapping. This has also been roughly applied to training recurrent network models \citep{wiseman2017training} but still requires unrolling the computation graph along the length of the sequence. Target propagation also requires a few things to work, notably that every layer in the network model must be an autoencoder and that a linear correction term is used to account for the imperfectness of auto-encoders (which can obstruct learning).

It is important to note that target propagation permits the use of non-linearities that output discrete values (or Bernoulli sampled activations). The TNCN also possesses this useful property, since the element-wise functions used to compute neuronal post-activations no longer need to be differentiable. This allows us to work with highly non-linear transformations where the gradients are often near zero, for example, stochastic binary units.

\subsection{Local Learning \& Greedy Layerwise Training}
\label{locality}
The desire for useful local learning, of which target propagation represents a strong modern step towards, is not new, and saw a small resurgence in the early days of training deeper networks in the form of layer-wise training of unsupervised models \citep{bengio_greedy_2007}, supervised models \citep{lee_deeply-supervised_2014}, and semi-supervised models \cite{ororbia_deep_hybrid_2015a,ororbia2015online} (also known as hybrid training).  However, among the many problems with these early approaches to deep learning was the lack of global coordination. Global coordination means that higher-level layers essentially direct lower-level layers in what patterns they should be extracting. A lower-level feature detector might be able to find different aspects of structure in its input since multiple patterns might satisfy its layer-wise objective. However, this detector will only locate the right bit of structure needed for the whole model to make sense, at any time-step, if a higher-level layer signals what pattern it should be looking for. Since greedy layer-wise approaches build the model from the bottom-up, freezing the learned lower-level parameters, this coordination is impossible to achieve.

The TNCN's localized learning approach was also motivated by the simple \emph{Bottom-Up-Top-Down} learning algorithm \citep{ororbia_deep_hybrid_2015b} , which showed that a stack of Boltzmann network modules (and other simple, auto-associative variations) could be learned in a ``pseudo-joint'' layerwise fashion. However, in order to build in some form of global coordination, a variation of back-propagation of errors was used, ultimately creating a global feedback path as part of the overall learning procedure. A more global approach was later presented in \cite{ororbia2015online}, incorporating top-down information much like that in \citep{salakhutdinov_efficient_2010}, however, these algorithms were only built for and studied on stateless problems. Furthermore, these approaches would be difficult to scale when extended to sequential modeling problems given their strong dependence on Markov Chain Monte Carlo sampling.

\subsection{Predictive Coding}
\label{pc}
Predictive coding theories posit that the brain is in a continuous process of creating and updating hypotheses that predict the sensory input it receives, directly influencing conscious experience \citep{panichello2013predictive}.
Models of sparse coding \citep{olshausen1997sparse} and predictive coding \citep{rao1999predictive} embody the idea that the brain is a directed generative model where the processes of generation (top-down mechanisms) and inference (bottom-up mechanisms) are intertwined \citep{rauss2013predictive} and interact to perform a sort of iterative inference of latent variables/states. Furthermore, nesting the ideas of predictive coding within the Kalman Filter framework \cite{rao1997dynamic} can create dynamic models that handle time-varying data. Many variations and implementations of predictive coding have been developed \citep{chalasani2013deep,lotter2016deep,santana2017exploiting}. Some approaches, such as predictive sparse decomposition \citep{kavukcuoglu2010fast}, attempt to speed up the iterative inference by introducing an inference network, but this again, creates a problem similar to that of variational inference--the generative model is constrained by the quality of the approximate inference model.

One key concept behind predictive coding that our own work embodies is that, for a multi-level objective to work well, each layer of a neural architecture would need an error feedback mechanism that could communicate the needs of the layer below it. If the learning signals are moved closer to the layers themselves, the error connections can directly transmit the information to the right representation units. Very importantly, this allows us to side-step the vanishing gradient problem that plagues pure back-propagation, where the internal layers of the architecture are trying to satisfy an objective that they only indirectly influence.
If we were to compare the updates from this local learning approach to back-propagation, the updates would still ascend/descend towards a similar objective, just not the steepest ascent/descent, so long as they were within 90 degrees of the direction given by back-propagation. However, since steepest ascent/descent is a greedy form of optimization, updates from a more localized approach might lead to superior generalization results \citep{broeke_thesis_2016}.

Our TNCN takes an adaptive, state-corrective approach similar to the dynamic predictive coding model proposed by \textcite{rao1997dynamic}. The general idea is to let the model first make a prediction and generate what it thinks the current frame or symbol will be at time $t$. The errors are computed for each layer via some feedback mechanism, starting from the sensors, which have direct access to the state of the world, and used to correct the model's internal states before moving on to predict the next time-step. A learning signal is created by comparing the corrected states to the initially predicted states. 
Since intra-layer competition among neuronal units is important (for reasons we will describe later), we also follow in the spirit of \textcite{olshausen1997sparse} and encourage sparsity indirectly through a penalty/constraint.

We combine the basic ideas described above to propose an algorithm for learning a temporal neural model, the Temporal Neural Coding Network. Generally, proposed alternative algorithms are tested on stateless data classification problems, such as the well known MNIST digit recognition dataset. In contrast we investigate the potential of our algorithm on sequential/temporal problems. Our approach can be viewed as on online, adaptive approach to learning, requiring no unrolling does back-propagation through time, since the TNCN is continuously engaged in self-correction, or rather, minimizing its total discrepancy between its expectations and targets. 

\section{Learning a Temporal Neural Coding Network}
\label{algo:dr}

Let us begin by formally defining a TNCN, at time $t$, with three layers of neural variables $(\mathbf{z}^0_t,\mathbf{z}^1_t,\mathbf{z}^2_t)$, where $\mathbf{z}^0_t$ refers to the output sensors that directly connect the model with the environment/world. The TNCN architecture distinguishes between two sets of parameters, $\Theta_g = \{G_2,G_1,V_2,V_1\}$ (the generative/predictive parameters) and $\Theta_e = \{E_2,E_1\}$ (the fixed, error feedback parameters). We define $(\widehat{\mathbf{h}}^0_t,\widehat{\mathbf{h}}^1_t,\widehat{\mathbf{h}}^2_t)$ to be the pre-activations of each latent variable.

To calculate the necessary statistics for one step of error correction, we first define the model's generated prediction for any (internal) arbitrary layer to be:
\begin{align}
\widehat{\mathbf{h}}^l_t &= V_l \mathbf{z}^l_{t-1} + G_{l+1} \mathbf{z}^{l+1}_t,\ 
\mathbf{z}^l_t = \phi^g(\widehat{\mathbf{h}}^l_t)
\end{align}
which assumes that any pre-activation is a linear combination of a filtration and a top-down (expectation) bias. The error units do two things: 1) create a local representation target by using information from the error units of the layer below and the TNCN's initial guess of the representation, 2) measure the discrepancy between this target and the corresponding layer pre-activation. Specifically, an error unit at layer $l$ within the model is computed as follows:
\begin{align}
\mathbf{e}^l_t = (\mathbf{h}^l_t - \widehat{\mathbf{h}}^l_t), \mbox{where, }\ \mathbf{h}^l_t = \phi^e(\widehat{\mathbf{h}}^l_t + \beta E_l \mathbf{e}^{l-1}_t)
\end{align}
noting that the second formula depicts how the error units compute a latent target using the error feedback weights $E_l$. $\phi^g(\cdot)$ is the post-activation function applied at each layer and $\phi^e(\cdot)$ is the element-wise function\footnote{In this paper, we used the hyperbolic tangent, scaled optimally according to \textcite{lecun1989generalization}.}  applied to the information coming from the error units below, meaning that the representation target is a non-linear function of the representation guess and the error fed back from below. This equation is reminiscent of the single corrective step found in dynamic predictive coding models formulated as Kalman Filters \citep{rao1997dynamic}. Note that these error feedback weights are fixed, as those of \textcite{lillicrap2014random} were, but they do not necessarily have to be.\footnote{Future work shall investigate neuro-biologically grounded ways of evolving these feedback weights.} However, the feedback weights of \textcite{lillicrap2014random} were used to carry the partial derivatives across layers (much like a short-circuit) and ultimately serve as a global feedback path, whereas the weights we propose are meant to help correct or update the currently guessed representation (which itself is a function of past information and the top-down generative weights of the layer above) and create local targets useful for subsequent learning. It is important to note that the error weights transmit error information \emph{behind} the non-linear activation function $\phi(\cdot)$ for any layer $l$. Once the correct pre-activation target $\mathbf{h}^l_t$ for any layer has been calculated, the final corrected representation is simply a re-application of the post-activation function for that layer, $\mathbf{z}^l_t = \phi(\mathbf{h}^l_t)$ (this will then be used as the vector summary of the past when moving on to the next time step $t+1$). A critical advantage of this proposed way of wiring the feedback connections (which is a unique departure from standard predictive coding models) is that we are now free to use any differentiable or non-differentiable non-linearity we like. This can include other sampling operations not amenable to the re-parametrization trick as well as discrete-valued activation functions (such as the classical hard threshold function).

We describe the TNCN's learning procedure, Discrepancy Reduction, next.

\subsection{The Discrepancy Reduction Algorithm}
\label{learn_algo}
To compute the gradients of model parameters once latent representations have been corrected, we exploit the local learning signals that natural arise given the way the error units we have designed work. The cost function that measures the total discrepancy within a TNCN composed of $l$ layers of latent variables, applied to real-valued input distributions, can be naively formulated as follows:
\begin{align}
\mathcal{L}(\mathbf{x}_t; \Theta_g, \Theta_e) = \sum_j \frac{-(\mathbf{x}_t[j] - \mathbf{z}^0_t[j])^2}{2 \sigma_x^2[j]} + \sum^L_{l=1} \sum_j \frac{-(\mathbf{h}^l_t[j] - \widehat{\mathbf{h}}^l_t[j])^2}{2 \sigma_l^2[j]} \label{loss}
\end{align}
noting that $\sigma^2_0 = \sigma^2_x = 1$. We fix the variance $\sigma^2_l = 1$, however, these can be additional parameters to be learned (details will appear in the appendix).  If we want to take an information-theoretic view, which aligns better with the idea of reducing internal discrepancy in the system, we can instead use the Kullback-Leibler Divergence \cite{kullback1951information} to measure the similarity between the guess and target for each local representation:
\begin{align}
\mathcal{L}(\mathbf{x}_t; \Theta_g, \Theta_e) &= \sum_j \frac{-(\mathbf{x}_t[j] - \mathbf{z}^0_t[j])^2}{2 \sigma_x^2[j]} - \sum^L_{l=1} KL(p_{\mathbf{h}^l_t} || q_{\widehat{\mathbf{h}}^l_t}) \\
\mathcal{L}(\mathbf{x}_t; \Theta_g, \Theta_e) &= \sum_j \frac{-(\mathbf{x}_t - \mathbf{z}^0_t)^2}{2 \sigma_x^2} - \sum^L_{l=1} \sum_j \Big ( \log  \frac{\sigma_{\widehat{\mathbf{h}}^l_t}[j]}{\sigma_{\mathbf{h}^l_t}[j]} + \frac{\sigma^2_{\mathbf{h}^l_t}[j] + (\mathbf{h}^l_t[j] - \widehat{\mathbf{h}}^l_t[j])^2}{2 \sigma^2_{\widehat{\mathbf{h}}^l_t}[j]} - \frac{1}{2} \Big ) \label{kl_loss}
\end{align}
noting that $\sigma^2_{\mathbf{h}^l_t}$ is the variance of $\mathbf{h}^l_t$ and $\sigma^2_{\widehat{\mathbf{h}}^l_t}$ is the variance of $\widehat{\mathbf{h}}^l_t$ (both of these are diagonal covariance matrices, and fixing these to vectors of ones further simplifies the expression to look quite similar to Equation \ref{loss}) . The leftmost term of the right-hand side of the equations for the loss is the partially grounded term in input space, while the rest of the terms are the higher-level terms (albeit simple and perhaps crude). Note that, internally, the TNCN architecture will readily compute the representation pre-activity targets $\{ \mathbf{h}^1_t, \mathbf{h}^2_t \}$ each time a sequential element $\mathbf{x}_t$ is presented.

\begin{figure*}
\centering
\begin{subfigure}{.5\textwidth}
  \centering
  \includegraphics[width=1.05\linewidth]{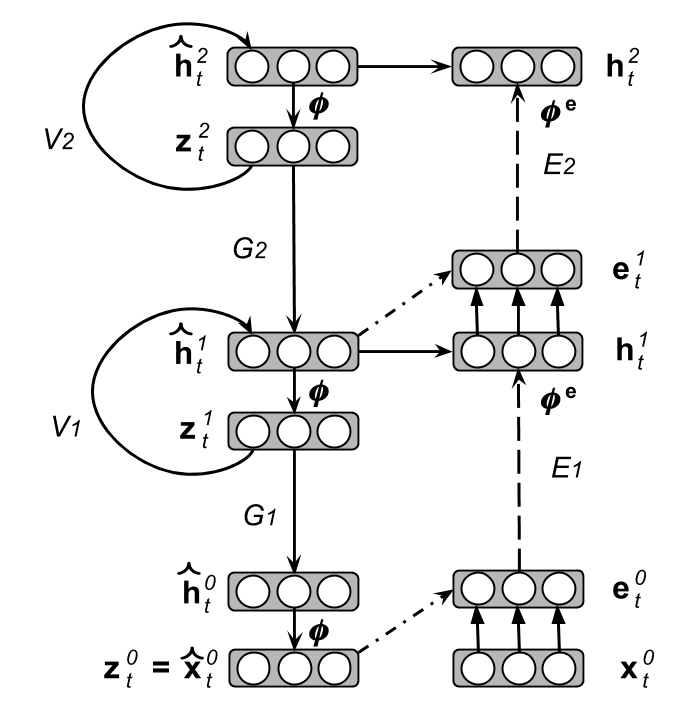}
  \caption{A two-layer TNCN architecture.}
  \label{fig:tncn_arch}
\end{subfigure}%
\begin{subfigure}{.5\textwidth}
  \centering
  \vspace{2.6cm}
  \includegraphics[width=1\linewidth]{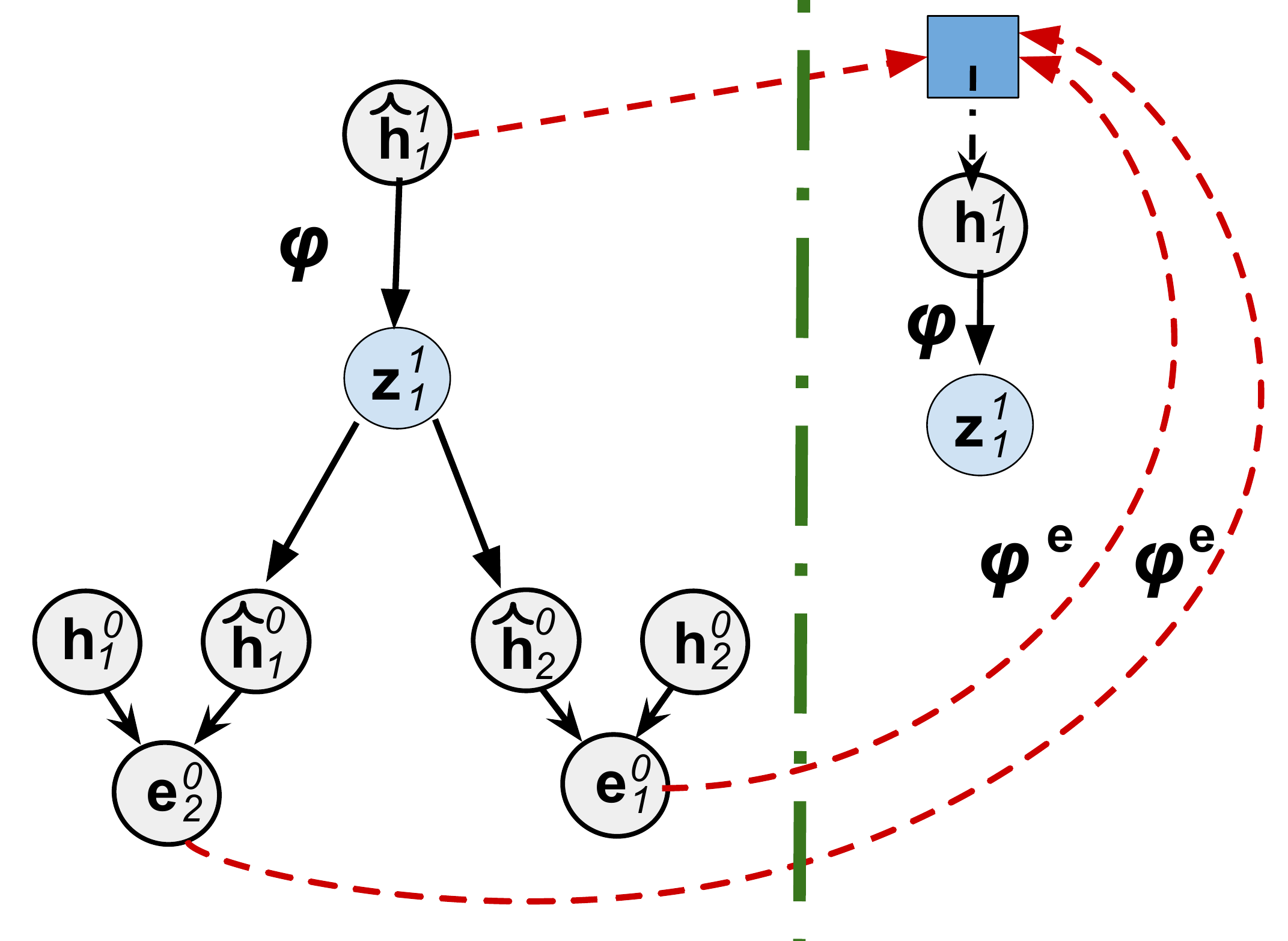}
  \caption{The feedback structure of a pair of cells.}
  \label{fig:tncn_cell}
\end{subfigure}
\caption{
In (\ref{fig:tncn_cell}), the error feedback loops are explicitly shown, where one pair of cells (in layer 0) communicate the discrepancy backwards to an earlier pair of cells (in layer 1). The error in layer 0 drives the representation target for layer 1, communicating the amount of mismatch in its own respective layer to help correct the representation above. In this sub-figure, solid lines represent the initial, feedforward pass of information while dashed lines represent the information flow along the structural feedback connections, which feed into an error unit that computes the target (which is also used to compute the corrected state). The dash-dotted line marks the separation of the forward phase and the error-correction phase in computing the latent variable $\mathbf{z}^1_1$. }
\label{fig:arch}
\end{figure*}

If we find the first-order partial derivatives of Equation \ref{loss} with respect to the weights in $\Theta_g$, we get the following updates (assuming we use stochastic gradient ascent as the update rule):
\begin{align}
\bigtriangledown_{G_{l+1}} &\propto \frac{\partial \mathcal{L}(\mathbf{x}_t, \Theta_g, \Theta_e)}{\partial G_{l+1}} \approx (\mathbf{h}^l_t - \widehat{\mathbf{h}}^l_t) (\mathbf{z}^{l+1}_t)^T + \eta_{G_{l+1}} \label{grad_G} \\
\bigtriangledown_{V_l} &\propto \frac{\partial \mathcal{L}(\mathbf{x}_t, \Theta_g, \Theta_e)}{\partial V_l} \approx (\mathbf{h}^l_t - \widehat{\mathbf{h}}^l_t) (\mathbf{z}^l_{t-1})^T + \eta_{V_l} \label{grad_V}
\end{align}
where $\eta$ is a noise process that is directly applied to the estimated parameter gradient. Such a process can be zero-mean Gaussian noise (with a scalar variance chosen through cross-validation). Note that the input layer $l = 0$ does not sport any recurrent connections (but if it did, these could be considered auto-regressive connections that relate input variables to past input variables) and is simply defined as:
\begin{align}
\bigtriangledown_{G_1} &\propto \frac{\partial \mathcal{L}(\mathbf{x}_t, \Theta_g, \Theta_e)}{\partial G_{1}} \approx (\mathbf{x}_t - \widehat{\mathbf{x}}_t) (\mathbf{z}^{1}_t)^T + \eta_{G_{1}},\ \bigtriangledown_{V_0} = 0 \label{out_grads}
\end{align}
where we simply set the gradient $\Delta_{V_0} = 0$ since setting the parameter matrix $V_0 = 0$ will effectively delete these recurrent input connections (we did this simply to avoid the situation where the dimensionality of the input is large, which would require even larger parameter matrices). One favorable property of the Discrepancy Reduction learning algorithm for learning TNCNs is the (partial) parallelism one may exploit in calculating parameter gradients, much like the goal of \citep{jaderberg2016decoupled}. One simply needs to run the TNCN's generation and target calculation procedures to get the guesses and targets, but once these statistics have been computed one can treat each layer as a computation sub-graph, the gradient estimates of which are independent of any other layer (this means that we design each layer to be more intricate and complex than what is used in this paper).

\begin{algorithm} 
\begin{algorithmic}
\State \textbf{Input:} $\mathbf{x}_t$ (mini-batch at time $t$), current parameters $\Theta_{t-1} = \{\Theta^g_t$, $\Theta^e_t\}$, previous state variables $\mathbf{Z} = \{\mathbf{z}^0_{t-1}, \mathbf{z}^1_{t-1}, \mathbf{z}^2_{t-1}\}$, and meta-parameters $\lambda$, $\beta$, $K > 0$

\Function{generate}{$\mathbf{Z}$, $\Theta^g_t$}
	\State $(\mathbf{z}^0_{t-1}, \mathbf{z}^1_{t-1}, \mathbf{z}^2_{t-1}) \leftarrow \mathbf{Z}$, $(G_1, V_1, G_2, V_2) \leftarrow \Theta^g_t$ \Comment Extract states and parameters
	\State $\widehat{\mathbf{h}}^2_t = V_2 \mathbf{z}^2_{t-1}$, $\mathbf{z}^2_t = \phi^g(\widehat{\mathbf{h}}^2_t)$ 
	\State $\widehat{\mathbf{h}}^1_t = V_1 \mathbf{z}^1_{t-1} + G_2 \mathbf{z}^2_t$, $\mathbf{z}^1_t = \phi^g(\widehat{\mathbf{h}}^1_t)$ 
	\State $\widehat{\mathbf{h}}^0_t = G_1 \mathbf{z}^1_t$, $\mathbf{z}^0_t = \phi^g(\widehat{\mathbf{h}}^0_t)$ 
	\State \Return $\widehat{\mathbf{h}}^0_t,\widehat{\mathbf{h}}^1_t,\widehat{\mathbf{h}}^2_t,\mathbf{z}^0_t,\mathbf{z}^1_t,\mathbf{z}^2_t$
\EndFunction

\Function{correct}{$\mathbf{x}_t,\mathbf{z}^0_t,(\widehat{\mathbf{h}}^0_t,\widehat{\mathbf{h}}^1_t,\widehat{\mathbf{h}}^2_t), \Theta^e_t, \beta, K$}
    \State $E_1, E_2 \leftarrow \Theta^e_t$
    \State $\mathbf{e}^0_t = \frac{\partial \mathcal{L}(\mathbf{x}_t,\mathbf{z}^0_t)}{\partial \widehat{\mathbf{h}}^0_t}$, $\mathbf{z}^0_{t+1} = \mathbf{x}_t$ \Comment If $K = 1$, set to datum at $t$
    \State $\mathbf{h}^1_t = \phi^e(\widehat{\mathbf{h}}^1_t + \beta E_1 \mathbf{e}^{0}_t)$, $\mathbf{e}^1_t = (\mathbf{h}^1_t - \widehat{\mathbf{h}}^1_t)$, $\mathbf{z}^1_{t+1} = \phi(\mathbf{h}^1_t)$ 
    \State $\mathbf{h}^2_t = \phi^e(\widehat{\mathbf{h}}^2_t + \beta E_2 \mathbf{e}^{1}_t)$, $\mathbf{e}^2_t = (\mathbf{h}^2_t - \widehat{\mathbf{h}}^2_t)$, $\mathbf{z}^2_{t+1} = \phi(\mathbf{h}^1_t)$ 
    \For{$k = 2$ to $K$}  \Comment Could add convergence criterion based on error units 
        \State $\widehat{\mathbf{h}}^1_t = V_1 \mathbf{z}^1_{t-1} + G_2 \mathbf{z}^2_{t+1}$ 
        \State $\widehat{\mathbf{h}}^0_t = G_1 \mathbf{z}^1_{t+1}$, $\mathbf{z}^0_{t+1} = \phi^g(\widehat{\mathbf{h}}^0_t)$
        \State $\mathbf{e}^0_t = \frac{\partial \mathcal{L}(\mathbf{x}_t,\mathbf{z}^0_{t+1})}{\partial \widehat{\mathbf{h}}^0_t)}$
    	\State $\mathbf{h}^1_t = \phi^e(\widehat{\mathbf{h}}^1_t + \beta E_1 \mathbf{e}^{0}_t)$, $\mathbf{e}^1_t = (\mathbf{h}^1_t - \widehat{\mathbf{h}}^1_t)$, $\mathbf{z}^1_{t+1} = \phi^g(\mathbf{h}^1_t)$ 
    	\State $\mathbf{h}^2_t = \phi^e(\widehat{\mathbf{h}}^2_t + \beta E_2 \mathbf{e}^{1}_t)$, $\mathbf{e}^2_t = (\mathbf{h}^2_t - \widehat{\mathbf{h}}^2_t)$, $\mathbf{z}^2_{t+1} = \phi^g(\mathbf{h}^1_t)$ 
    \EndFor
    \State \Return $\mathbf{e}^0_t,\mathbf{e}^1_t,\mathbf{e}^2_t,\mathbf{h}^1_t,\mathbf{h}^2_t,\mathbf{z}^0_{t+1},\mathbf{z}^1_{t+1},\mathbf{z}^2_{t+1}$
\EndFunction

\Function{updateModel}{$\mathbf{x}_t, \mathbf{Z}, \Theta^g_t, \Theta^e_t, \beta, K, \lambda$}
	\LineComment 1) Run generative model (get guesses of each representation)
	\State $\widehat{\mathbf{h}}^0_t,\widehat{\mathbf{h}}^1_t,\widehat{\mathbf{h}}^2_t,\mathbf{z}^0_t,\mathbf{z}^1_t,\mathbf{z}^2_t \leftarrow \Call{generate}{\mathbf{Z}, \Theta^g_t}$ 
    \LineComment 2) Correct states via error-created targets
	\State $\mathbf{e}^0_t,\mathbf{e}^1_t,\mathbf{e}^2_t,\mathbf{h}^1_t,\mathbf{h}^2_t,\mathbf{z}^0_{t+1},\mathbf{z}^1_{t+1},\mathbf{z}^2_{t+1} \leftarrow \Call{correct}{\mathbf{x}_t,\mathbf{z}^0_t,(\widehat{\mathbf{h}}^0_t,\widehat{\mathbf{h}}^1_t,\widehat{\mathbf{h}}^2_t), \Theta^e_t, \beta, K}$ 
    \LineComment 3) Adjust parameters via gradient ascent \& output corrected state variables
    \State $(G_1, V_1, G_2, V_2) \leftarrow \Theta^g_t$  \Comment Extract current parameters
    \State $\bigtriangledown_{G_1} = \mathbf{e}^0_t (\mathbf{z}^{1}_t)^T $, $\bigtriangledown_{V_1} = \mathbf{e}^1_t (\mathbf{z}^{1}_{t-1})^T + \frac{\partial \Omega(\widehat{\mathbf{h}}^1_t)}{\partial V_1}$
    \State $\bigtriangledown_{G_2} = \mathbf{e}^1_t (\mathbf{z}^{2}_t)^T + \frac{\partial \Omega(\widehat{\mathbf{h}}^1_t)}{\partial G_2}$, 
    $\bigtriangledown_{V_2} = \mathbf{e}^2_t (\mathbf{z}^{2}_{t-1})^T + \frac{\partial \Omega(\widehat{\mathbf{h}}^2_t)}{\partial V_2}$
    \State $\tilde{G}_1 = G_1 + (\lambda \bigtriangledown_{G_1} + \eta_{G_1})$, $\tilde{V}_1 = V_1 + (\lambda \bigtriangledown_{V_1} + \eta_{V_1})$
    \State $\tilde{G}_2 = G_2 + (\lambda \bigtriangledown_{G_2} + \eta_{G_2})$, $\tilde{V}_2 = V_2 + (\lambda \bigtriangledown_{V_2} + \eta_{V_2})$
    \State $ \Theta^g_t = \{ \tilde{G}_1,\tilde{V}_1,\tilde{G}_2,\tilde{V}_2 \}$, $\mathbf{Z} = \{ \mathbf{z}^0_{t+1},\mathbf{z}^1_{t+1},\mathbf{z}^2_{t+1} \}$
    \State \Return $\Theta^g_t, \mathbf{Z}$
\EndFunction

\end{algorithmic}
\caption{The Discrepancy Reduction learning algorithm for building TNCN model with two latent variable layers. $\Omega( \widehat{\mathbf{h}}^l_t )$ is the regularization function (or prior distribution) imposed on a layer's pre-activities, $\widehat{\mathbf{h}}^l_t$.}
\label{discrepancy_reduction}
\end{algorithm}

In predictive coding theories of the brain, it is often assumed that sparsity is a key ingredient. This means we seek representations of data where only a small subset of the latent variables have non-zero values. If the TNCN is to disentangle concepts in its latent representations, the need for sparsity makes sense since it is reasonable to assume that only a few out of the many possible concepts/variables explain any given datum (useful in tasks such as corrupt image denoising \citep{cho2013simple}). From a theoretical perspective on feature learning, we desire compact representations of the input in which no information is lost regarding the input \citep{bengio2013deep,van2015theory}. Dense representations, though rich, are highly entangled since small changes in the input can lead to big changes in the representation vector. Sparse representations, in contrast, are robust and mostly conserve the set of non-zero features \citep{glorot2011deep}. In this paper, we only enforce a ``weak'' form of lateral competition over the neuronal variables through a simple Laplacian prior distribution over the pre-activities. During the learning phase/step, lateral competition patterns (where the neurons fight to be active) get internalized in the model parameters via the term $\Omega(\widehat{\mathbf{h}}^l_t)$.\footnote{It is important to note that the simple way we encourage sparsity does not mean that all representations of the TNCN are guaranteed to be sparse. It is quite possible that the model could produce dense representations for data points outside the training sample since only during  training is sparsity encouraged. One way to remedy this would be ensure that the Laplacian prior is active during inference (as in classical sparse coding). To better encourage sparsity, other prior distributions, such as the spike-and-slab distribution \citep{goodfellow2012spike} (to avoid controlling the  pre-activation magnitudes), or architectural modifications \citep{szlam2011structured} might improve generalization and will be the subject of future work.} We found in initial preliminary experiments that sparsity was indeed a necessary component in improving performance.

With all of the above taken together, the full algorithm for the TNCN (generation, representation correction, and parameter updating) is depicted in Algorithm \ref{algo:dr}. The TNCN (or rather its generation/inference mechanisms) and its learning algorithm, are intricately tied together, since the learning procedure will make use of the representation targets created by the architecture's error-driven correction mechanism. This operates in the spirit of predictive coding which posits that the brain's generation and inference procedures interact to formulate local learning signals. The mechanism we use for target creation is rather simple, and future work should investigate more sophisticated mechanisms (especially ones with evolving error weights). As is depicted in Algorithm \ref{algo:dr}, the representation-correction mechanism can be extended to a process where targets can be iteratively refined, in the hopes of shortening the overall learning phase. 

With respect to higher-level objectives, we can see that the error units play a crucial role--they are in fact the first-order derivatives of the Gaussian log likelihood (with fixed unit variance). Learning is simple since the error units can be easily re-used to calculate parameter gradients incrementally (when combined with the competition prior) and the only activation function derivative required in this approach is that of the output distribution model (which can be easily worked out for commonly used output distributions, such as the Gaussian, Bernoulli, and Categorical distributions). Note that better error units could be derived if one chose a different tactic for measuring the distance between predicted and corrected representation layers (for example, one could measure the Manhattan distance instead of the Euclidean distance, as formulated in our framework). However, the general idea is that the TNCN is engaged with ensuring its layer-wise representations are as close to those suggested by the error units--it is optimizing not only on the input space, but also in the latent space giving us some rough measure of the quality of the model's internal representations. In some sense, this bears a loose resemblance to the \emph{bottom-up-top-down} algorithm proposed in \citep{ororbia_deep_hybrid_2015b}, which proposed a non-greedy way of learning a set of layer-wise experts. Through the feedback mechanism and the top-down generation paths, the local learning rules of the TNCN gain some form of global coordination, which was lacking in the greedy approaches of the past \citep{bengio_greedy_2007,ororbia_deep_hybrid_2015a} when training deep belief networks and their hybrid variants.

It is important to highlight that learning and inference under this model is ideally supposed to be continuous, meaning that the model simultaneously generates expectations and then corrects itself (both representations and parameters) each time a new datum from a sequence is presented. This makes the model directly suited to learning incrementally from data-streams. Furthermore, the TNCN shows how two types of recurrence/feedback are at play when modeling sequences:  1) the model is recurrent across the temporal axis since it is stateful, since each processing layer depends on a vector summary of the past, and 2) the model is structurally recurrent, similar to deep Boltzmann machines and Hopfield Networks \citep{hopfield1982neural}, since error is fed back in order to automatically correct guessed representations.

\section{Experimental Results}
\label{experiments}

\subsection{The Bouncing Balls Process}
\label{exp:bballs}
To test our proposed TNCN architecture and its learning algorithm, Discrepancy Reduction, we benchmark our performance on the bouncing balls dataset following in line with the experimental setup used in \cite{sutskever2009recurrent}. This high-dimensional dataset was created by simulating the rudimentary physics of three balls bouncing around in a box. We generate a training set of 4000 training sequences and a test set of 200 sequences (as well as yet another 200 sequences to create a development set). Furthermore, our models are given no prior knowledge of the task, e.g. convolutional weight matrices, much as was done in \cite{taylor2007modeling}. On this dataset, \emph{Frame t-1} is the simplest possible baseline--a model that predicts the next step as simply the previously seen frame.

We trained TNCNs with two and three layers of latent variables, searching for the size of the layers over the range $\{1000-3000\}$ (with performance measured on the validation subset). In this experiment, the logistic sigmoid activation function ultimately proved to be the most useful (we also experimented with nonlinearities, but found these to not work as well). Parameters were initialized from centered Gaussian distributions with $\sigma^2 = 0.01$ (except in the three-layer TNCN, setting the top-level recurrent and generative weights using $\sigma^2 = 0.025$ improved performance a bit). Error feedback parameters were initialized with centered Gaussians of $\sigma^2 = 0.1$ (again, in the case of the 3-layered model, we used $\sigma^2 = 0.025$ for the top layer). The value of the gradient noise was set to $\sigma_g = 0.005$ (to control the stochastic approximation of the prior over weights). $k$ was to $1$. \footnote{We also experimented with a naive approach by fixing $k > 1$ and found that performance slightly worsened. We believe that to properly use iterative inference, we would need employ an adaptive iterative inference schedule, since in the early states of learning, the model is learning to use its error feedback weights, but in later stages one should raise $k$ gradually to give the model the opportunity to iteratively refine its representations.} Gradients (at each time step) were estimated using mini-batches of 50 samples (across 50 parallel videos). Parameters were updated using the method of steepest gradient descent, of which we employed the \emph{Adam} \citep{kingma2014adam} adaptive learning rate scheme with the step-size fixed to $\lambda = 0.0001$. We further apply norm-rescaling to the gradients computed by Discrepancy Reduction (threshold is $15$) \citep{pascanu2013difficulty} and take the Polyak average \citep{polyak1992acceleration} of the model at its best performance on the validation subset (i.e., early stopping).

We report our models' average squared next-step prediction (20 trials) per frame in Table \ref{results:bballs} and compare against previously reported errors. The proposed TNCN performs better than the Boltzmann-based models.
Furthermore, we see that the inclusion of an additional hidden layer actually helps the directed model, pushing it to nearly the same level as a deep temporal sigmoid belief network. Note that all of the models we compare our TNCN to utilize back-propagation through time as a core mechanism while our approach is incremental and adaptive. To improve the performance of our model, we believe using an adaptive iterative inference scheme combined with learnable variance parameters are key ingredients.

What is most surprising is that our simple way of building non-linear error units was effective in creating useful local representation targets. This is evidenced by the fact that performance vastly improves upon adding a layer of these types of neurons to the top-down generative model. What this might mean is that the TNCN is making use of the generated local targets and trying to minimize the mismatch between its initial guess of the representation (conditioned on past corrected representations) and the error-corrected representation. Since each layer higher up in the network aims to do a better job at explaining the layer representation below, this local target becomes useful during learning and inference. The target in effect helps keep the model on track as it updates its latent representations given the sequence data it encounters, step by step.

\begin{table*}[!t]
\renewcommand{\arraystretch}{0.485}
\caption{Test-set performance on the bouncing ball problem.}
\centering
\label{results:bballs}
\centering
\begin{tabular}{ll}

\textbf{Ball Model}&\textbf{Error} \\
\hline 
\emph{Frame t-1} & $11.73 \pm 0.28$ \tabularnewline
\emph{TSBN-1} \parencite{gan2015deep} & $9.48 \pm 0.39$\tabularnewline
\emph{RTRBM} \parencite{sutskever2009recurrent} & $3.88 \pm 0.33$ \tabularnewline
\emph{SRTRBM} \parencite{mittelman2014structured} & $3.31 \pm 0.33$ \tabularnewline
\emph{TSBN-4} \parencite{gan2015deep} & $3.07 \pm 0.40$ 
\tabularnewline
\emph{DTSBN-D} \parencite{gan2015deep} & $2.99 \pm 0.42$ \tabularnewline
\emph{DTSBN-s} \parencite{gan2015deep} & $2.79 \pm 0.39$ \tabularnewline
\hline

\emph{2-TNCN} (present work)  & $3.55 \pm 0.19$ \tabularnewline %
\emph{3-TNCN} (present work)  & $3.03 \pm 0.14$ \tabularnewline

\end{tabular}
\end{table*}

\section{Conclusions}
\label{conclusion}
In this paper, we proposed a novel neural architecture, the Temporal Neural Coding Network (TNCN), and its learning algorithm, Discrepancy Reduction. To derive our idea, we drew inspiration from several strands of work that seek biologically-plausible alternative learning algorithms that generalize better to out-of-sample data. Furthermore, we connected these various research paths with the aim of tackling the difficult problem of learning representations of data in an unsupervised manner. With this target in mind, we argued for the use of higher-level objectives, which can be interpreted as local learning rules that still result in a globally coherent model. We developed our model with the goal of learning from streams of sequential data, inspired by the dynamic variations of predictive coding theories of the brain. On one generative modeling benchmark, the bouncing balls problem, we showed that our model, using an incremental, adaptive procedure, can compete with various models that use back-propagation through time.

Breaking free from the global feedback path required in back-propagation, as we do in this work, brings us closer to building models that are better suited for true unsupervised learning \citep{barlow1989unsupervised}. Unsupervised learning requires the computer to capture \emph{all} possible dependencies between \emph{all} observed variables, since inputs are no longer distinguished from outputs, as they in supervised learning. It is this latter form of learning that is more closely related to how humans learn, where much of the incoming data does not come with labels (or, at best, comes with very few labels, as mentioned in \cite{ororbia_deep_hybrid_2015a,ororbia_deep_hybrid_2015b}). A successful unsupervised learning system is one that can discover all of the useful concepts and underlying causes to explain what it perceives \citep{lecun2015deep} just as an infant must do, by observation alone. 

More importantly, learning generic representations in an unsupervised system would further free us from the rather inflexible models created from the task-specific nature of supervised learning. Since downstream supervised/reinforcement learning approaches focus on task-specific measurements, the objectives used in unsupervised learning must then attempt to measure the quality of the generic representations acquired by the generative models we train.  
Defining what a good-quality general representation is itself an open problem and an active area of theoretical research \citep{van2015theory,mcnamara2016modular}. However, in lieu of the ideal metric, we took a small step towards higher-level objectives by literally interpreting this concept as a set of simple reconstruction loss terms that measure how well our neural architecture can predict a set of representation targets. Far better performance can be reached, we hypothesize, if better representation measurements and metrics can be developed. 

We are continuing to run experiments on other datasets to show the generality of our architecture and learning algorithm, and ultimately seek to use the learned generative models in downstream supervised learning tasks. Furthermore, we believe that our incremental, adaptive algorithm is better suited to streaming problems, more commonly found in the online learning setting \citep{ororbia2015online}, which we argue is important when considering the even greater challenge of lifelong learning \citep{mitchell2015never}.

\section{Acknowledgements}
We would like to thank Yoshua Bengio for useful feedback.

\bibliographystyle{apa}  
\bibliography{ml}   


\end{document}